\documentclass[12pt]{article}

\usepackage{newtxtext,newtxmath}
\usepackage{relsize}
\usepackage{multirow}
\usepackage{graphicx}
\usepackage{bm}
\usepackage{amsmath}
\usepackage{mathtools}
\usepackage[letterpaper,margin=1in]{geometry}

\makeatletter
\newcommand{\Biggg}{\bBigg@{4}}
\newcommand{\biggg}{\bBigg@{2.5}}
\makeatother

\renewenvironment{abstract}
	{\quotation}
	{\endquotation}

\date{}

\makeatletter
\renewcommand{\fnum@figure}{\textbf{Figure \thefigure}}
\renewcommand{\fnum@table}{\textbf{Table \thetable}}
\makeatother

\usepackage{scicite}

\usepackage{url}

\def\scititle{
    Reconciling distributed compliance with high-performance control in continuum soft robotics
}
\title{\bfseries \boldmath \scititle}

\author{
	Vito Daniele Perfetta$^{\ast}$,
    Daniel Feliu-Talegon$^{\dagger}$,\and
    Ebrahim Shahabi$^{\dagger}$,
	Cosimo Della Santina\and
	\small Cognitive Robotics Department, Faculty of Mechanical Engineering, \\
    \small Delft University of Technology, Delft 2628 CD, The Netherlands.\and
	\small$^\ast$Corresponding author. Email: V.D.Perfetta@tudelft.nl\and
	\small$^\dagger$These authors contributed equally to this work.
}

\begin{document} 

\maketitle

\begin{abstract} \bfseries \boldmath

High-performance closed-loop control of truly soft continuum manipulators has remained elusive. Experimental demonstrations have largely relied on sufficiently stiff, piecewise architectures in which each actuated segment behaves as a distributed yet effectively rigid element, while deformation modes beyond simple bending are suppressed. This strategy simplifies modeling and control, but it sidesteps the intrinsic complexity of a fully compliant body and makes the system behave as a serial kinematic chain—much like a conventional articulated robot. An implicit conclusion has consequently emerged within the community: distributed softness and dynamic precision are incompatible.

Here we show that this trade-off is not fundamental. We present a highly compliant, fully continuum robotic arm—without hardware discretization or stiffness-based mode suppression—that achieves fast and precise task-space convergence under dynamic conditions. The platform integrates direct-drive actuation, a tendon routing scheme enabling coupled bending and twisting, and a structured nonlinear control architecture grounded in reduced-order strain modeling of underactuated mechanical systems. Modeling, actuation, and control are co-designed to preserve essential mechanical complexity while enabling high-bandwidth loop closure.

Experiments demonstrate accurate and repeatable execution of dynamic Cartesian tasks, including fast positioning and interaction.
The proposed system achieves the fastest reported task execution speed for continuum soft manipulators. 
At millimetric precision, execution speed increases by nearly fourfold compared with prior approaches—while operating on a fully compliant continuum body. These results demonstrate that distributed compliance and high-performance dynamic control can coexist, opening a path toward truly soft manipulators that approach the operational capabilities of rigid robots without sacrificing morphological richness.

\end{abstract}

\noindent

\section*{INTRODUCTION}


A central motivation fueling the development of the thriving soft robotics field is the ambition to realize systems that exploit distributed compliance to generate rich and adaptive deformation patterns~\cite{rus2015design,whitesides2018soft,xie2023octopus,yue2025embodying}. While this vision has profoundly shaped materials and fabrication strategies~\cite{laschi2016soft,wang2024sensing,bang2024bioinspired}, control has followed a more conservative trajectory. To date, although impressive, experimental demonstrations of high-performance control~\cite{thuruthel2018model,santina_planarSR2020,centurelli2022,fischer2023dynamic,bruder2025koopman,azizkhani2025dynamic,Rus_soft2026} have largely focused on soft manipulators built as series interconnections of actuated segments that are sufficiently stiff to remain weakly affected by internal dynamic coupling and gravitational loading. Consequently, their deformation is predominantly confined to bending along circular arcs, or to similarly predictable modes. Each segment thus behaves effectively as a rigid element governed by simplified bending coordinates, naturally aligning with control formulations that assume full actuation.
This structural bias is not merely a matter of mechanical convenience. Theoretical tractability has long relied on restricting the deformation space and suppressing modes that are difficult to stabilize—or even to represent analytically. Mechanical design and control have therefore co-evolved in a self-reinforcing loop: piecewise architectures make control feasible, and control assumptions legitimize piecewise architectures. As a consequence, experimental demonstrations of soft robotic intelligence remain largely confined to systems that closely resemble fully actuated rigid kinematic chains. In doing so, they constrain the expressive mechanics of a genuinely compliant continuum body and fall short of the very original motivations of soft robotics.

This tension leads to a central question: \textit{Can a general continuum arm achieve high-speed and accurate task-space regulation without relying on structural simplifications?} Addressing it requires confronting distributed elasticity, intrinsic underactuation, and configuration-dependent equilibrium constraints simultaneously, rather than abstracting them away. Recent theoretical research has increasingly embraced this challenge by engaging explicitly with the high-dimensional nature of continuum morphology. Variable-strain formulations provide finite-dimensional dynamic representations grounded in continuum mechanics~\cite{li2023piecewise,mathew_reduced_2025}. Structured Lagrangian control of underactuated systems~\cite{della2023model,Pustina_coll2024} and closed-loop inverse kinematics mappings~\cite{santina_clik25} offer systematic tools for task-space regulation in mechanically coupled settings. Although often investigated in isolation and validated primarily in simulation, these results suggest that the perceived limitation may not be fundamental. Still, an integrated control architecture combining all these elements and capable of handling fully soft robots remains missing; both as a methodological result and, most importantly, as an experimental reality.

In this work, we address this gap by showing that, when integrated within a single functioning architecture, recent advances in continuum modeling and underactuated control are not only able to manage complex deformations in fully soft robots, but can also push their performance beyond that reported for segmented soft manipulators. To this end, mechatronic design, modeling, and control are developed jointly to preserve mechanical expressivity while ensuring stable and responsive dynamics. We introduce a continuum soft arm explicitly conceived to maintain distributed compliance along its entire body, avoiding segmented construction and stiffness-dominated simplifications. A tendon-routing architecture enables coupled bending and twisting deformations, allowing configuration patterns that extend beyond pure curvature modes. The actuation pipeline is engineered for transparent loop closure, enabling operation beyond quasi-static assumptions.
Control is achieved through a novel framework grounded in variable-strain continuum modeling of underactuated mechanical systems. A reduced-order Lagrangian formulation captures the essential elastic dynamics while remaining computationally tractable in real time. Task-space regulation is implemented via a closed-loop inverse kinematics mapping that explicitly accounts for equilibrium constraints induced by underactuation. A low-level nonlinear controller ensures rapid convergence toward the shape prescribed by the task-space layer.

Experiments validate the proposed architecture across positioning and dynamic tasks. 
The system achieves the fastest reported task execution speed for continuum soft manipulators. At millimetric steady-state precision, execution speed increases by nearly fourfold compared with prior approaches (Table~\ref{tab:state_of_art_comparison}), while operating on a fully compliant continuum body. The proposed structure thus takes a decisive step away from rigid-like morphologies while narrowing the performance gap with them, demonstrating that distributed compliance and high-performance dynamic control can coexist.

\newcommand{\CircLow}{$\bullet\,\circ \circ\,\, \circ$}
\newcommand{\CircMedium}{$\bullet\,\bullet \circ\,\, \circ$}
\newcommand{\CircHigh}{$\bullet\,\bullet \bullet\,\, \circ$}
\newcommand{\CircVeryHigh}{$\bullet\,\bullet \bullet\,\, \bullet$}

\newcommand{\SquareLow}{$\blacksquare\square\square\square$ }
\newcommand{\SquareMedium}{$\blacksquare\blacksquare\square\square$ }
\newcommand{\SquareHigh}{$\blacksquare\blacksquare\blacksquare\square$ }
\newcommand{\SquareVeryHigh}{$\blacksquare\blacksquare\blacksquare\blacksquare$ }

\newcommand{\SquareLowThree}{$\blacksquare\square\square$ }
\newcommand{\SquareMediumThree}{$\blacksquare\blacksquare\square$ }
\newcommand{\SquareHighThree}{$\blacksquare\blacksquare\blacksquare$ }
\newcommand{\SquareVeryHighThree}{$\blacksquare\blacksquare\blacksquare$ }

\newcommand{\ArrowLow}{$\uparrow\textcolor{gray}{\uparrow}\textcolor{gray}{\uparrow}\textcolor{gray}{\uparrow}$}
\newcommand{\ArrowMedium}{$\uparrow\,\,\uparrow\textcolor{gray}{\uparrow}\,\,\textcolor{gray}{\uparrow}$}
\newcommand{\ArrowHigh}{$\uparrow\,\,\uparrow\,\,\uparrow\textcolor{gray}{\uparrow}$}
\newcommand{\ArrowVeryHigh}{$\uparrow\uparrow\uparrow\uparrow$}

\begin{table} 
	\centering
	\caption{\textbf{Comparison of the proposed work with the most relevant experimental results on tas 
    k-space dynamic control of soft robots.} {\small 
    Results are organized for segmented, non-segmented reduced-compliance bending, and general continuum soft structures, reporting deformation strains, material softness, model and control techniques$^\dagger$, generalizability, preliminary training requirements, stability guarantee (with $\sim$ indicating limited validity), task-level feedback, 3D task achievement, and the Root Mean Square of the Task-Space error. Finally, task velocity is given for millimeter-level (errors~$<1$~cm) and centimeter-level (errors~$<3$~cm) precision scenarios, where dashed lines indicate unmet performance criteria and $(\cdot)^*$ denotes trajectory reference velocity when point-to-point regulation is absent and actual speed values are not reported.
    Additional details are reported in the Supplementary Material.
  } 
   }
	\label{tab:state_of_art_comparison} 
    \scriptsize

\resizebox{\columnwidth}{!}{%
\begin{tabular}{l|l|ccccc|cc|c|}
\hline
 & & \multicolumn{5}{c|}{\textbf{Segmented}} & \multicolumn{2}{c|}{\textbf{Non-segmented Bending}}& \multicolumn{1}{c|}{\textbf{General}} \\
&  & \cite{azizkhani2025dynamic} & \cite{centurelli2022} & \cite{Rus_soft2026} & \cite{santina_planarSR2020} & \cite{fischer2023dynamic} & \cite{haggerty2023control} & \cite{bruder2020data} & This Work \\
\hline
\multirow{7}{*}{\rotatebox[origin=c]{90}{\textbf{Robot}}} & $\uparrow$ Material softness &  \SquareMedium  & \SquareMedium & \SquareMedium  & \SquareHigh & \SquareHigh &  \SquareLow & \SquareMedium & \SquareVeryHigh \\
 & Deformation strains &  &  &  &  &  &  &  &   \\
 & ~~~~Torsion & $\times$ & $\times$ & $\times$ & $\times$ & $\times$ & $\times$ & $\times$ & \checkmark \\
 & ~~~~Bending & \checkmark & \checkmark & \checkmark & \checkmark & \checkmark & \checkmark & \checkmark & \checkmark \\
 & ~~~~Axial & \checkmark  & \checkmark & \checkmark & \checkmark & $\times$ & $\times$ & $\times$ & \checkmark \\
 & 3D Motions & \checkmark & \checkmark & $\times$ & $\times$ & \checkmark & \checkmark & $\times$ & \checkmark \\
\hline
\hline
\multirow{6}{*}{\rotatebox[origin=c]{90}{\textbf{Model \& Control}}} & Model & PCC & LSTM & DNN + R & PCC & PCC & KO & KO & VS \\
 & Technique & AP & TRPO & N-I L & CS & PD+ & K-LQR & MPC & CLIK + PID+ \\
\cline{2-10}
 & $\uparrow$ Generalizability & \SquareLowThree & \SquareMediumThree & \SquareHighThree & \SquareLowThree & \SquareLowThree & \SquareMediumThree & \SquareMediumThree & \SquareHighThree \\
 & No Training Required & \checkmark  & $\times$ & $\times$ & \checkmark & \checkmark & $\times$ & $\times$ & \checkmark \\
 & Stability Guaranteed & $\sim$ & $\times$ & $\sim$ & $\sim$ & $\sim$ & $\sim$ & $\sim$ & \checkmark \\
 & Task-level Feedback & $\times$ & \checkmark & \checkmark & \checkmark & \checkmark & \checkmark & $\times$ & \checkmark \\
\hline
\hline
\multirow{3}{*}{\rotatebox[origin=c]{90}{\textbf{Results}}} 
 & $\uparrow$ cm-task speed [m/s] & $\textbf{0.42}^*$ & $0.05^*$ & $0.019$ & $0.028$ & $0.060$ & $0.37$ & $0.006$ & $\underline{0.384}$ \\
 & $\uparrow$ mm-task speed [m/s] & $-$ & $-$ & $\underline{0.017}$ & $-$ & $-$ & $-$ & $0.006$ & $\textbf{0.063}$ \\
 & $\downarrow$ Task-space RMS error [mm] & $12.91$ & $7.47$ & $\underline{0.81}$ & $11.24$ & $14.09$ & $25.54$ & $2.31$ & $\textbf{0.67}$ \\
\hline
\end{tabular}
}
 \begin{minipage}{\columnwidth}
\footnotetext[0]{\footnotesize $^\dagger$VS, Variable Strain; KO, Koopman Operators; PCC, Piecewise Constant Curvature; LSTM, Long-Short Term Memory; DNN+R, Deep Neural-Network and Reptile; CLIK\,+\,PID+, Closed Loop Inverse Kinematics and Collocated PID with gravity and elasticity compensation; K-LQR, Koopman--Linear Quadratic Regulator; PD+, PD with acceleration potential field; AP, Adaptive Passivity; CS, Cartesian Stiffness; TRPO, Trust Region Policy Optimization; N-I\,L, Neuron-Inspired Learning; MPC, Model Predictive Control.}
\end{minipage}
\end{table}


\begin{figure}
	\centering
	\includegraphics[width=1\textwidth]{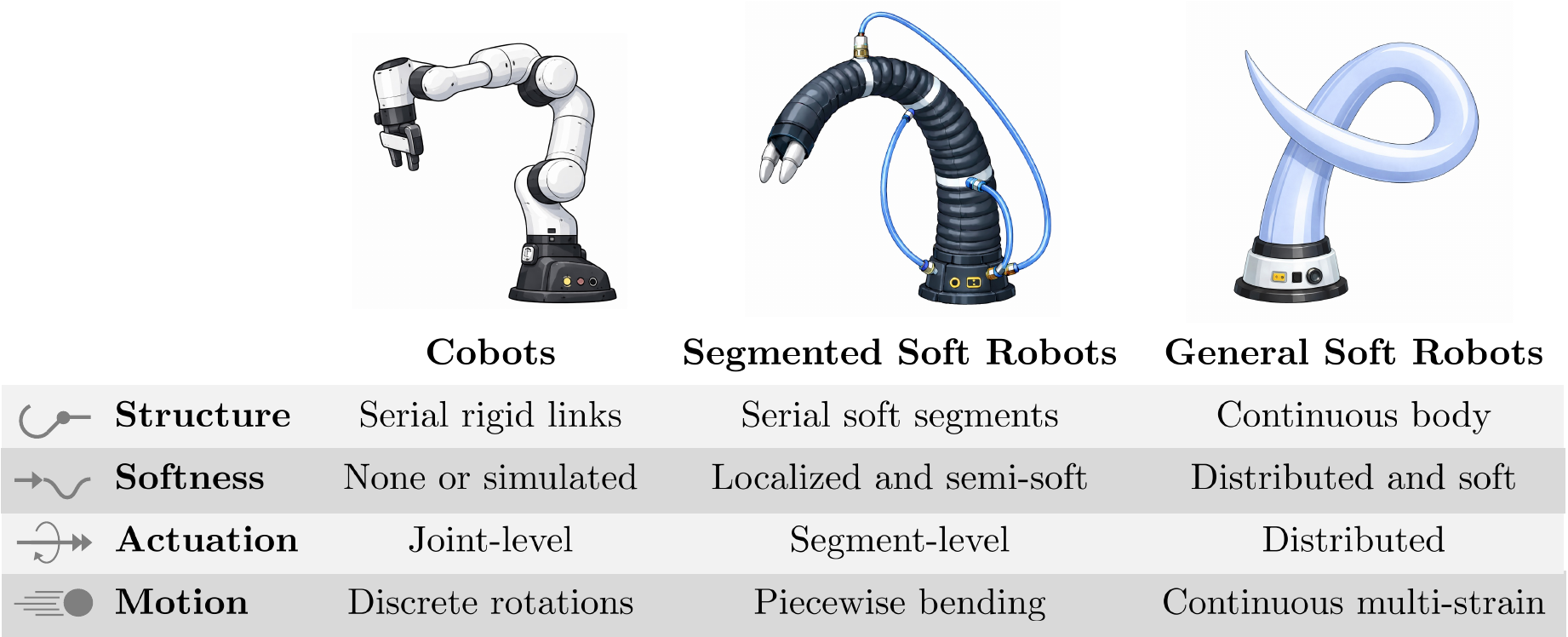} 
	\caption{\textbf{From Cobots to Soft Robots.} Comparison of physical characteristics across collaborative rigid robots, segmented soft robots, and general soft robots. The table highlights key differences in structure, material softness, actuation, and motion capabilities.}
	\label{fig:graphical_abstract} 
\end{figure}

\section*{RESULTS}

The objective of the experimental evaluation is to assess whether explicitly integrating theoretical results that have so far remained largely confined to analytical investigations (reduced-order strain modeling, nonlinear underactuated dynamics, and model-aware control of continuum systems) can enable a purposefully designed fully compliant continuum arm to substantially outperform the segmented architectures commonly studied in the soft robotics literature and to move toward the dynamic performance traditionally associated with rigid manipulators. 

Industrial collaborative robots routinely achieve Cartesian speed in the order of $0.2$--$0.3$~m~s$^{-1}$ in certified safe human--robot interaction modes~\cite{franka_research3_datasheet_2025}, with precision in the order of $3$--$8$~mm range~\cite{franzesecal}. State-of-the-art segmented soft manipulators, by contrast, operate at markedly lower speeds, especially when millimeter accuracy is required - see Table~\ref{tab:state_of_art_comparison}. The experiments that follow are designed to precisely position the proposed system within this landscape and quantify what becomes achievable when contemporary continuum modeling and underactuated control tools are instantiated on a truly soft body. 

The first set of results isolates the mechatronic and control-theoretic contributions. The continuum soft manipulator is actuated via high-performance direct-drive motors and a distributed tendon-routing architecture, enabling coupled bending and twisting along the entire body. Motion is regulated by a nested nonlinear control framework operating in task space.

We then move to quantitative results. Task-space regulation is assessed in two- and four-dimensional obstacle-free scenarios. For each target, a fixed time window is defined and progressively reduced, starting from a duration long enough to guarantee convergence to millimetric accuracy (see Materials and Methods). This procedure identifies the minimum time required for consistent task completion as speed demands increase.

To enable comparison with prior work, two Cartesian error thresholds are considered (Table~\ref{tab:state_of_art_comparison}): a centimetric band (error $< 3$~cm per relevant dimension) and a millimetric band (error $< 1$~cm). For each threshold, we report an equivalent task-accomplishment velocity, defined as tip displacement divided by the time required to reach and remain within the band. The error vector is computed as $e_{x}=x_\mathrm{d} - x$, where $x$ and $x_\mathrm{d}$ are the actual and desired task-space coordinates, defined according to the task-dimension considered.

Performance is further validated on representative real-world tasks, including sugar pouring and pendulum striking. 
Across all experiments, instantaneous tip velocities on the order of meters per second are observed alongside millimetric steady-state accuracy - a regime typically associated with rigid manipulators, here obtained with a fully compliant continuum structure.

\subsection*{Multi-modal actuation in a fully continuum soft robot} 

In this work, we present a highly deformable soft robotic arm capable of achieving a rich, complex workspace through a novel tendon routing strategy that simultaneously enables bending and twisting within a single continuous elastomeric body.

\begin{figure}
	\centering
	\includegraphics[width=0.68\textwidth]{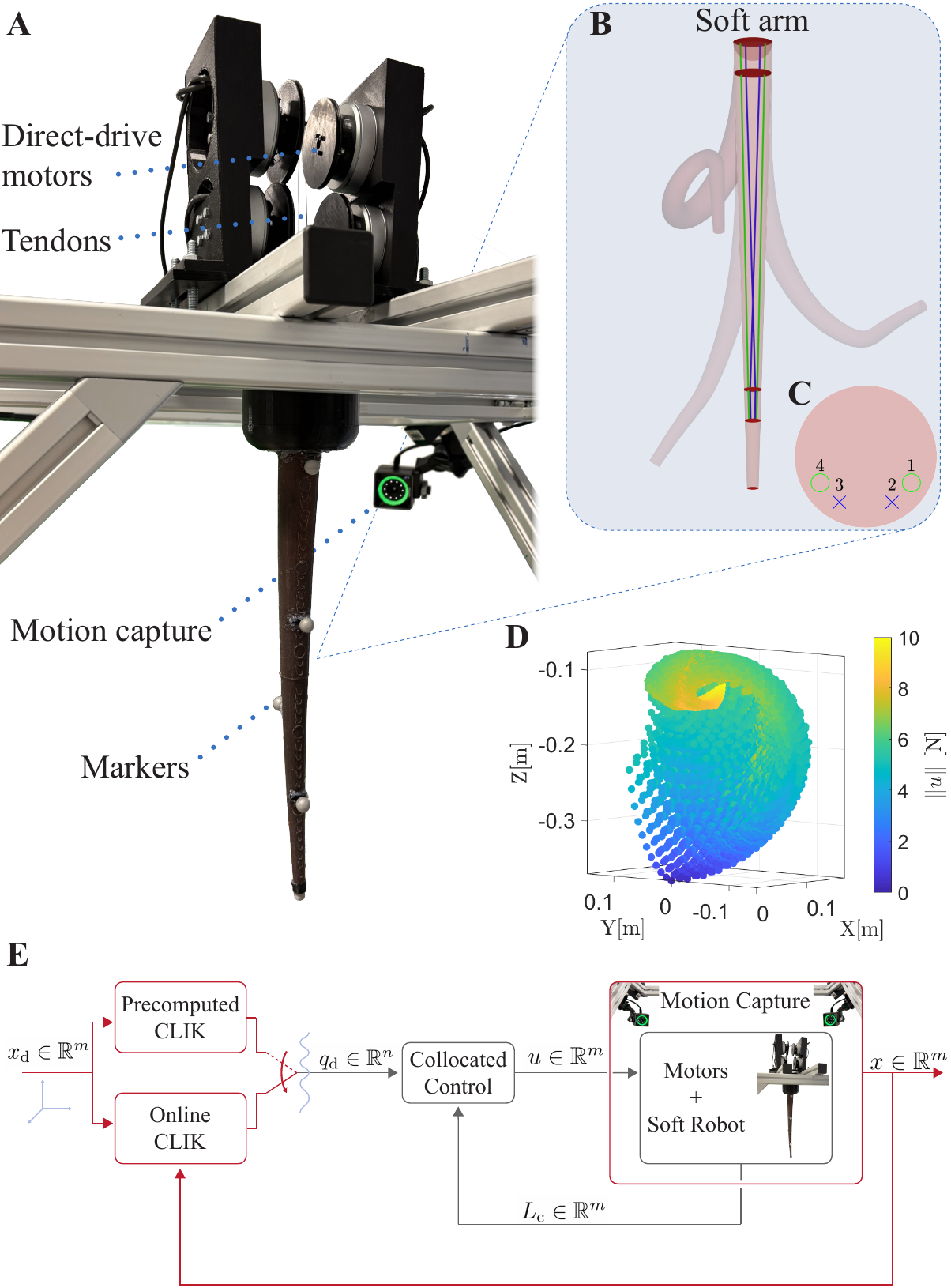} 
    	\caption{\textbf{Experimental Setup and Control Framework.} (\textbf{A}) Experimental setup. 
        (\textbf{B}) Internal view of the soft arm. Different tendon configurations are visually distinguished by straight (green) and crossed (blue) paths, while burgundy circles denote the tiny guiding rings. Transparent renderings on the sides illustrate examples of achievable shapes. (\textbf{C}) Top view of the robot illustrating the tendons' placement at the base. Each tendon–motor pair is identified by a number from 1 to 4, with green circles denoting straight tendon paths and blue crosses denoting the crossed ones. (\textbf{D}) Soft robot's workspace attained through combined actuation of the four tendons with pulling forces in~$\left[0,-5\right]$~N. (\textbf{E}) Block diagram of the two-level control framework employed, with the inner loop in gray and the outer loop in red. }
	\label{fig:setup} 
\end{figure}

The arm is fabricated as a single $38$~cm-long elastomeric structure using Ecoflex 00-50 (Smooth-On, USA), shaped into a cone with base radius $R_\mathrm{b} = 1.60$~cm and tip radius $R_\mathrm{t} = 0.48$~cm. Four tendons are embedded during casting and distributed along the body, occupying approximately half of the arm's vertical cross-section in two distinct groups (Fig.~\ref{fig:setup}B). Two tendons run straight along the lateral sides at angles of $30^\circ$ and $150^\circ$ from the cross-sectional horizontal axis (Fig.~\ref{fig:setup}C), extending $32.5$~cm from the base, while the remaining two are positioned in the frontal region at $60^\circ$ and $120^\circ$ in a crossed configuration over the same length.

This mixed straight–crossed tendon layout intentionally breaks axial symmetry, unlocking both decoupled and combined bending and twisting modes within a single soft body, without relying on helical reinforcements, segmented joints, or layered material architectures. Actuating the straight lateral tendons primarily produces bending, whereas actuating the crossed frontal tendons induces twisting about the longitudinal axis. Simultaneous activation of both groups generates coupled bending–twisting motions, and when multiple tendons are engaged, the arm also undergoes limited axial compression due to its inherent softness. The result is a hemispherical workspace that curves onto itself near the tip, enabling configuration patterns that extend well beyond pure curvature modes, as illustrated in Fig.~\ref{fig:setup}D.

To define the tendon path and minimize friction during actuation, each tendon was inserted into a thin silicone tube before casting. The tube-tendon assemblies were positioned in the mold using $2$~mm guide rings before the elastomer was poured and cured. Despite these measures, residual friction effects persist and are explicitly accounted for in the modeling framework presented in Material and Methods. Detailed fabrication steps are provided in the Supplementary Material.

\subsection*{A two-level control architecture for general soft robots} 
The architecture leverages and extends the latest advances in configuration-space control and CLIK for Lagrangian underactuated systems to friction-aware Variable Strain models, achieving fast and millimeter-level task-space accuracy on a highly underactuated soft robotic arm.

The architecture consists of a model-based, two-level nested control loop, illustrated in Fig.~\ref{fig:setup}E.
 The inner loop implements a configuration-space collocated controller that computes the tendon pulling forces $u \in \mathbb{R}^{n_\mathrm{a}}$ according to the error in the actuation coordinates $\left(\bar{\theta}_\mathrm{a,d}-\bar{\theta}_\mathrm{a}\right)\in \mathbb{R}^{n_\mathrm{a}}$, reconstructed from the measured tendon lengths $L_\mathrm{c} \in \mathbb{R}^{n_\mathrm{a}}$ (see Material and Methods). The outer loop maps a desired task-space pose $x_\mathrm{d} \in \mathbb{R}^m$ to a reference configuration $q_\mathrm{d}\in \mathbb{R}^{n}$ via a CLIK algorithm that feeds back the robot's actual pose. From $q_\mathrm{d}$, the desired tendon lengths $L_\mathrm{c,d}$ and the actuation coordinates $\bar{\theta}_\mathrm{a,d}$ are then analytically derived. This decomposition allows each level to address a distinct challenge: the inner loop handles the underactuated dynamics, while the outer loop compensates for modeling inaccuracies and small disturbances in task space, where the number of controlled task-space coordinates matches the number of available actuators, i.e. $n_\mathrm{a} = m$.

Two variants of the CLIK algorithm are employed. The first is a precomputed approach that relies exclusively on model-based tip position predictions and can therefore be executed offline, enabling fast motions without real-time computational overhead. However, due to unmodeled dynamics and parameter uncertainties, this approach alone yields task-space errors of a few centimeters. The second is an online implementation that incorporates feedback from the robot's actual pose $x$, measured via a motion-capture system tracking markers attached to the arm, reducing the task-space error to the millimeter range at the cost of increased computational demand.  

To exploit the complementary strengths of both variants, the experimental protocol consists of first driving the robot toward the desired pose $x_\mathrm{d}$ using the precomputed CLIK, and subsequently refining convergence by switching to the real-time version, unless stated otherwise.

\subsection*{Free-space regulation tasks}
\subsubsection*{Free-space coiling motion}
\begin{figure}
	\centering
	\includegraphics[width=0.9\textwidth]{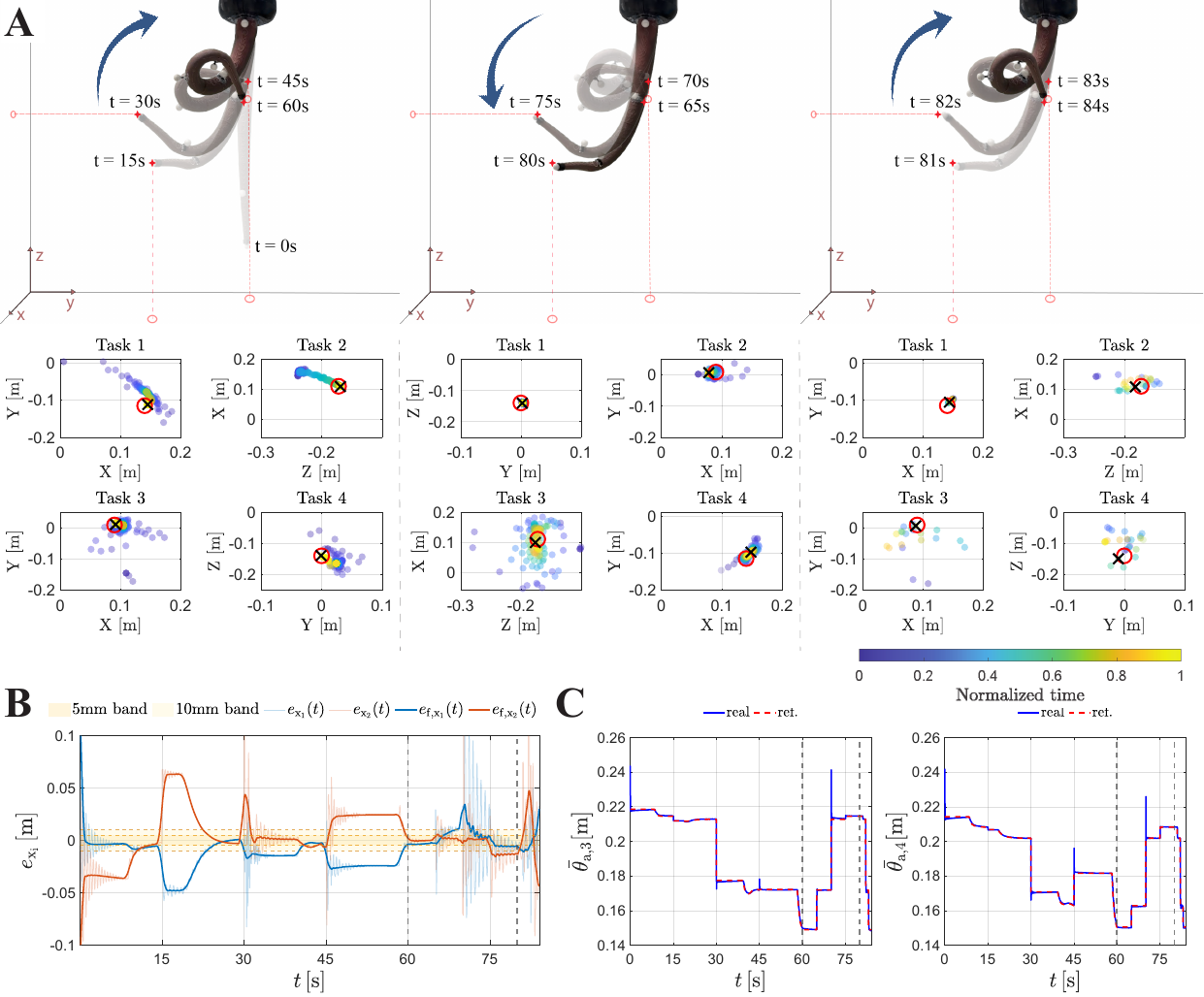} 
	\caption{\textbf{Two-dimensional coiling task.} (\textbf{A}) (Top) The robot’s final configurations at each reference Cartesian coordinate pair are shown. Reference points are marked with red circles, while the corresponding 3D projections reached by the robot are highlighted with red stars. From right to left, the shapes correspond to time windows of~$15$~s,~$5$~s and~$1$~s, respectively, with a transparency gradient representing progression over time. (Bottom) The projection of the robot’s motion onto the Cartesian plane is shown for each task and time interval. Red circles denote desired target coordinates, while black crosses indicate the robot’s final position before reference change. Motion is reported on a normalized time scale. (\textbf{B}) Time evolution of the errors in operational space and their convergence within~$5$~mm and~$10$~mm bands. To improve readability, the post-processed error obtained through a moving-average filter is highlighted, while the raw error signal is displayed more transparently in the background. Grey dashed vertical bars indicate the switch in time windows.
    (\textbf{C}) Time evolution of the estimated actuation coordinates (blue) and their tracking of the reference signals (red) under configuration-space collocated control.}
	\label{fig:2m_coiling} 
\end{figure}

The soft manipulator is here regulated to reach predefined tip positions in a variable two-dimensional, obstacle-free task space, inducing a coiling motion that fully leverages the soft structure’s intrinsic deformability, as shown in Fig.~\ref{fig:2m_coiling} and movie~S1. To this end, only actuators 3 and 4 (see Fig.~\ref{fig:setup}C) are activated.

 Four pairs of reference coordinates for the tip position are assigned across the three Cartesian planes (see Fig.~\ref{fig:2m_coiling}A). The soft manipulator is commanded to sequentially achieve each coordinate pair, with transition times of~$15$~s during the upward motion. Once a desired point is reached, the corresponding configuration is stored and later provided as reference to the configuration-space controller to assess the manipulator's achievable speeds and repeatability during the subsequent downward motion, with transition times of~$5$~s, and again during the final upward motion lasting~$1$~s per point. 

From Fig.~\ref{fig:2m_coiling}B, it can be observed that steady-state errors always converge in the~$5$~mm band when the robot operates within the~$15$~s time window. However, when the transition duration is reduced to both~$5$~s and~$1$~s, the oscillations are not fully damped, resulting in slightly larger errors prior to each reference change that nonetheless mostly remain within the~$10$~mm band. These oscillations are mainly caused by the unactuated state variables, which are not directly controllable through the control inputs (see Material and Methods), whereas the actuated coordinates are accurately tracked by the low-level controller, as illustrated in Fig.~\ref{fig:2m_coiling}D. 
During the entire experiment, the soft arm reaches peak velocities of up to~$3.63$~m~s$^{-1}$, which typically occur when a new precomputed configuration-space reference is assigned following a desired pose change. Meanwhile, convergence to the desired configuration within a~$3$~cm error band along both task-space directions was achieved at an equivalent task velocity of up to~$0.189$~m~s$^{-1}$, whereas velocities of approximately~$0.02$~m~s$^{-1}$ were required to attain millimetric accuracy. 
\subsubsection*{Free-space triangle configuration}
\begin{figure}
	\centering
	\includegraphics[width=0.76\textwidth]{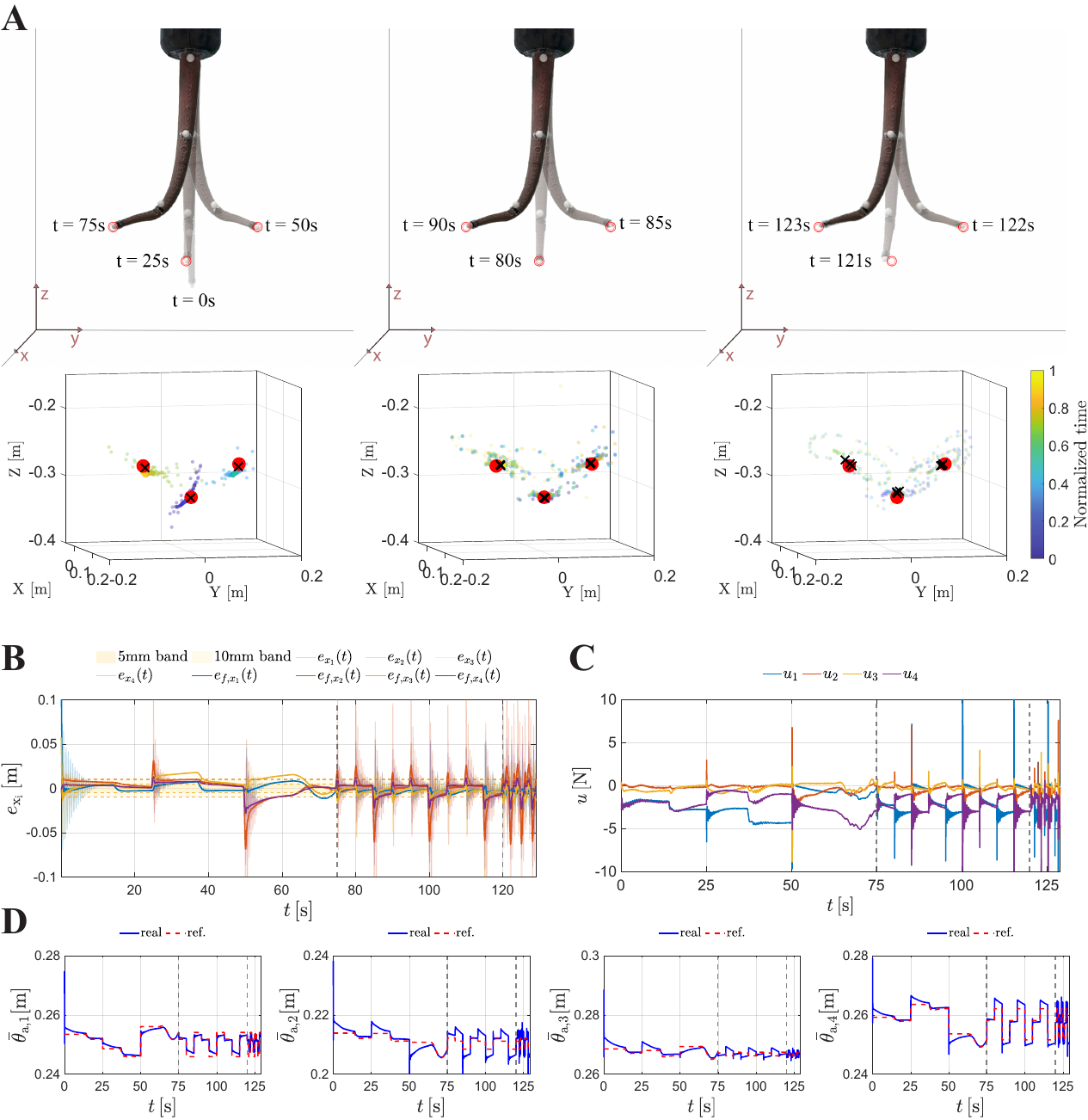} 
	\caption{\textbf{Four-dimensional triangular shape task.}  (\textbf{A}) (Top) The robot’s final configurations upon converging at reference points (red circles) displaced in a triangular shape in the Cartesian space are shown. From right to left, the shapes correspond to time windows of~$25$~s,~$5$~s and~$1$~s, respectively. (Bottom) The reconstructed robot’s motion in the Cartesian space is shown for each time interval. The red circle denotes the desired targets, while the black cross indicates the robot's last position before the reference change.  Motion is reported on a normalized time scale. (\textbf{B}) Time evolution of the errors in operational space and their convergence within the~$5$~mm and~$10$~mm accuracy bands. To improve readability, the post-processed error obtained through a moving-average filter is highlighted, while the raw error signal is displayed more transparently in the background. Grey dashed vertical bars indicate the switch in time windows. (\textbf{C}) Time evolution of the pulling forces exerted by the tendons. (\textbf{D}) Time evolution of the estimated actuation coordinates (blue) and their tracking of the reference signals (red) under configuration-space collocated control.}
	\label{fig:4m_triangle} 
\end{figure}

In this experiment, the scalability of the proposed control architecture is assessed in an extended four-dimensional obstacle-free workspace. In this setting, the tip must reach three target points arranged in a triangular configuration while progressively increasing its velocity, showcasing fast, precise and repeatable convergence to a configuration, as illustrated in Fig.~\ref{fig:4m_triangle} and movie~S2.  

Specifically, the CLIK algorithm controls four task-space coordinates: three regulate the tip position, while the fourth constrains the $y$-coordinate of the marker immediately preceding the tip in the global frame. The motion sequence proceeds in two phases. First, the robot is commanded to reach each of the three triangle vertices sequentially, with a $ 25$~s time window per target, during which the real-time CLIK converges to the desired pose and the resulting configuration is stored. In the second phase, the stored configurations are replayed to the low-level controller in sequence, repeating each transition three times with a time slot of $5$~s and then again three times with a time slot of $1$~s, to assess convergence speed and repeatability of the proposed framework.

 Each transition is repeated three times to assess the repeatability of the proposed framework.

The time-evolution of the corresponding task-space errors is shown in Fig.~\ref{fig:4m_triangle}B. Here, it can be noticed how the robot's tip accurately converges to each target point with an error norm below~$5$~mm during the initial phase with a~$25$~s time window. In the subsequent step, the error remains on the millimetric scale even when the transition time is reduced to~$5$~s, during which the tip velocity reaches peaks of approximately~$2.52$~m~s$^{-1}$. When the time window is further shortened to~$1$~s, the manipulator completes the transition with an average error still within the~$10$~mm band at each point, while instantaneous errors continue to oscillate and do not fully decay within the assigned time. In this case, convergence to the desired configuration within a~$3$~cm error band along all four task-space directions was achieved at an equivalent task velocity of up to~$0.384$~m~s$^{-1}$, whereas a speed of approximately~$0.063$~m~s$^{-1}$ was required to attain millimetric accuracy.

\subsection*{Application-driven tasks}
\subsubsection*{Dynamic ball interaction task}
\begin{figure}
	\centering
	\includegraphics[width=0.74\textwidth]{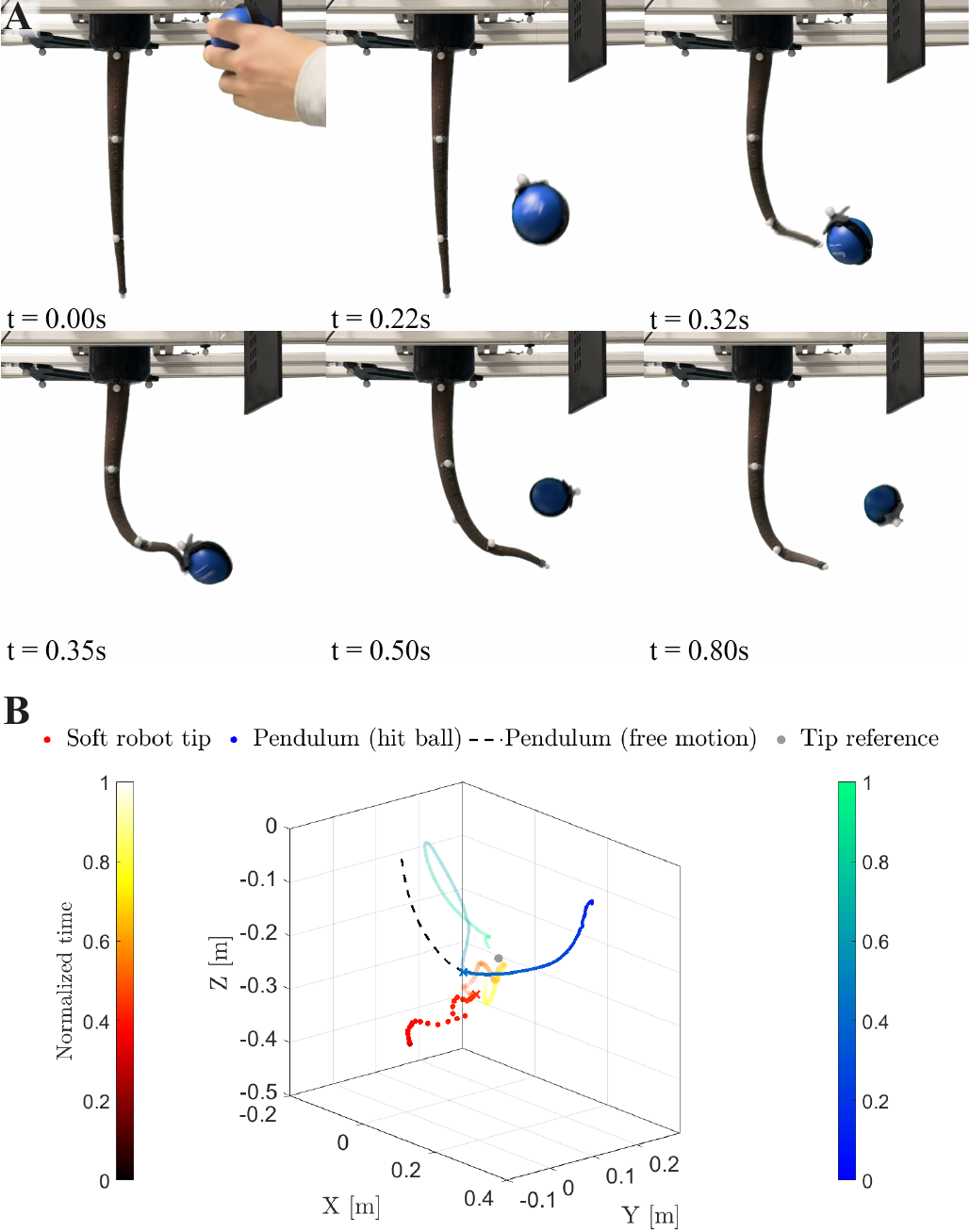} 
	\caption{\textbf{Pendulum-like ball striking.}
		(\textbf{A}) The time sequence of shapes attained by the soft manipulator is reported here. The ball is first released at $t=0.00$~s, and then accurately struck at $t=0.35$~s by the robot. (\textbf{B}) The motion of the robot tip together with the trajectory of the marker placed on the external surface of the pendulum is shown. The corresponding normalized time histories are reported on the left (soft robot) and on the right (pendulum). The trajectories include cross markers, shown in the corresponding colors, to identify the impact instant; post-impact trajectories are subsequently displayed with increased transparency. In addition, the desired target position for the robot's tip is indicated by the gray circle, while the nominal trajectory of the pendulum's marker in the absence of impact is depicted by the black dashed line.}
	\label{fig:4m_ball} 
\end{figure}

To demonstrate the high speeds and precision achievable with the proposed framework, the soft arm is commanded to strike a pendulum mass at the instant of maximum velocity, as shown in Fig.~\ref{fig:4m_ball} and movie~S3. 

The additional experimental setup consists of a mass of approximately~$21$~g suspended from the robot cage by a massless cable of length~$30.5$~cm, with an initial angular displacement of about~$80^\circ$~from the vertical. A 3D-printed pierced wall is placed on the side to ensure repeatability of the initial condition. 
Given the initial position of the pendulum pivot at
$\left(0.147,0.156,0.073\right)$~m, 
and the time evolution of the system, the pendulum's pose at its velocity peak can be estimated. This value is then used as the desired Cartesian target $x_\mathrm{d}$ for the CLIK algorithm, enabling interception by the soft arm.

In this experiment, the CLIK block is fully precomputed to minimize execution delays and synchronize the control action with the pendulum motion. As a result, the manipulator effectively strikes the ball at approximately~$t=0.35$~s (see Fig.~\ref{fig:4m_ball}A), achieving a peak speed of about~$4.37$~m~s$^{-1}$. The equivalent task velocity, here computed as the ratio of the difference between striking and initial tip poses over the corresponding time interval, reaches up to~$1.48$~m~s$^{-1}$. 
Figure~\ref{fig:4m_ball}B illustrates the resulting trajectory deviation of the marker attached to the pendulum from its nominal path after impact, characterized by a pronounced displacement along the $y$-axis.

\subsubsection*{Sugar manipulation task} 
\begin{figure}
	\centering
	\includegraphics[width=1\textwidth]{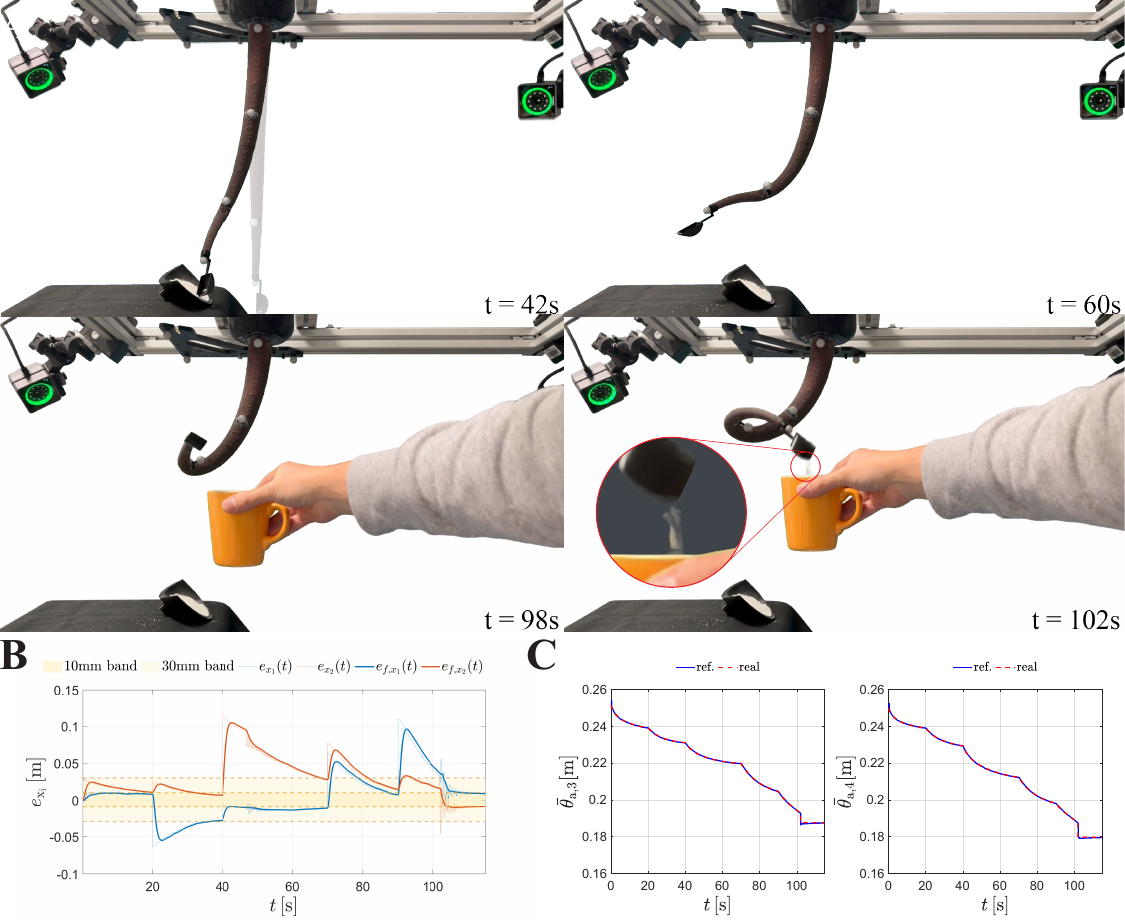} 
	\caption{\textbf{Sugar pick-and-pour task to sweeten a cup of coffee.} (\textbf{A}) The sequence of shapes attained by the soft manipulator is here reported for $t=42,60,98,102$~s. Specifically, the arm accesses the sugar container at $t=42$~s, and eventually releases it at $t=102$~s. The starting configuration of the robot is shown transparently in the top-left picture. (\textbf{B}) The time evolution of the errors in operational space is shown, together with their convergence within the~$10$~mm and~$30$~mm accuracy bands. To improve readability, the post-processed error obtained through a moving-average filter is highlighted, while the raw error signal is displayed more transparently in the background. 
    (\textbf{C}) The time evolution of the estimated actuation coordinates (blue) and their tracking of the reference signals (red) under configuration-space collocated control is shown.}
	\label{fig:2m_sugar} 
\end{figure}

In this experiment, six pairs of reference coordinates for the tip position are assigned across the three Cartesian planes to perform a pick-and-pour task, where sugar is delicately grasped from a container and released into an approaching cup of coffee. 
 This complex task showcases the manipulator’s shock-absorption capabilities and its ability to coordinate multiple deformations (i.e., axial, bending and torsion) to perform safe and precise motion, even during compliant interactions with the environment (see Fig.~\ref{fig:2m_sugar} and movie~S4), 
 demonstrating the control architecture’s robustness to small unmodeled dynamics and external disturbances thanks to the outer loop closure.

The sugar is placed in a 3D-printed container fixed on a support inside the manipulator's workspace. 
Two reference points are first defined before and inside the sugar container, allowing the~$5$~cm 3D-printed spoon attached to the robot’s tip to collect the sugar. Then,  three additional via points guide the manipulator along the desired trajectory, with a final point positioning the tip directly above the cup placed in front of the robot's base. To prevent sugar spillage, the robot moves at a slower speed, maintaining an average transition time of approximately~$20$~s between reference coordinates. The whole time evolution is illustrated in Fig.~\ref{fig:2m_sugar}A. 

In this experiment, the CLIK algorithm was executed entirely in real time, without any initial feedforward computation during the reference transitions. Only 
at approximately~$102$~s is the precomputed version employed to send the final target configuration to the low-level controller, specifically to provide sufficient thrust to release the sugar from the spoon.
As shown in Fig.~\ref{fig:2m_sugar}B, the control architecture effectively compensates for unmodeled external disturbances. Indeed, the steady-state error remains within the~$10$~mm and~$30$~mm bands for the first two reference points even under environmental contacts during sugar collection. Subsequently, although the robot converges more slowly toward the desired Cartesian coordinates due to the unmodeled increased tip weight from the spoon and added sugar, the controller continues to reduce the error toward the millimetric range. As a result, the task is successfully executed, even though steady-state errors remain at the centimetric scale because transitions to the next reference occur before full convergence.

\section*{DISCUSSION}
This work demonstrates that fully continuum soft robots can achieve high-speed, accurate task-space regulation without any structural segmentation or reduced-compliance body, with performance approaching that of rigid manipulators. 

Thanks to the novel actuation path, shapes that simultaneously combine bending and twisting can now be achieved, thereby enabling the execution of real-world tasks that are difficult or unattainable with existing approaches. 
Tendon-driven actuation is a practical and versatile choice for soft robotic systems, as it allows the motors to be located away from the deformable structure and the actuation paths to be reconfigured to tailor the robot's workspace. However, friction effects along the tendon paths introduce configuration-dependent behaviors that must be carefully accounted for in both fabrication and modeling to achieve precise control.
In this regard, the proposed friction modeling extension to the VS framework (see Material and Methods) significantly enhances accuracy, effectively capturing tendon–structure interactions that would otherwise lead to substantial modeling errors. 
To model and control the proposed soft robot, we have presented a fully interpretable framework. Here, the combination of precomputed CLIK with configuration-space collocated control enables both high instantaneous velocities and rapid convergence. This dynamic performance is reinforced by a real-time CLIK layer, which enhances robustness against discrepancies between the mathematical model and the physical manipulator, allowing steady-state Cartesian errors on the order of millimeters. Notably, relying on a fully model-based architecture not only provides clear insight into the system dynamics and allows analytical stability guarantees, but also enables a higher level of robustness and generalizability, allowing fast and precise motion across a variety of tasks with different deformations and operating conditions without any training procedure.

Given that rigid robots still outperform soft systems in terms of (sub-)millimetric accuracy and task execution speed, the proposed manipulator has nevertheless demonstrated the ability to attain instantaneous velocities on the same order as collaborative robots (up to approximately $4$~m~s$^{-1}$) and steady-state errors below~$5$~mm. In terms of task-achievement speed, our soft arm exceeds the previous best tracked performance (see Table~\ref{tab:state_of_art_comparison}), reaching a task-space error below $3$~cm at a velocity of $0.384$~m~s$^{-1}$ and convergence to a millimetric error at a speed approximately four times higher than the current state of the art, registering~$0.063$~m~s$^{-1}$~versus the previous~$0.017$~m~s$^{-1}$. 
Notably, these results not only advance beyond prior performance, but are also achieved using a softer and fully continuous manipulator undergoing much more complex deformations.

Overall, while the presented results demonstrate the potential of the proposed framework to extend the operational boundaries of soft robots toward real-world applications, several limitations and opportunities remain.  First, the experimental results reported in this study focus exclusively on regulation to desired configurations, as this is the operating condition under which the stability of the employed controllers has been formally established in prior works~\cite{Pustina_coll2024, santina_clik25}. Extending the framework to Cartesian trajectory tracking constitutes a natural next step and will be addressed in future works.
Second, the current implementation relies on a motion capture system to accurately estimate the shape of the soft manipulator. For real-world deployment, it will be necessary to overcome this dependency by integrating alternative sensing modalities. Finally, the proposed approach does not explicitly model interactions with the external environment, which can still influence performance, as observed in the sugar grasp-and-pour experiment. Explicitly accounting for environmental interaction effects within the control framework will be essential to further improve robustness and overall system performance in real-world scenarios.

\section*{MATERIAL AND METHODS}
Here, we describe the implementation of each block of the control architecture introduced in Fig.~\ref{fig:setup}E. We first present the Variable Strain modeling procedure, with a focus on the representation of friction effects. Next, we describe the configuration-space collocated controller used to regulate the robot to a desired reference shape, detailing the proposed actuation coordinates and their analytical relationship with the tendon lengths. Finally, we present the CLIK algorithm, which maps the desired tip pose to a reference configuration in the configuration space.

\subsection*{Dynamic model of the soft arm}
Continuum soft robots are typically characterized by slender geometries, which allow an effective reduced-order description of their dynamics through deformations of the central axis alone, as is the case for elastic rods. By exploiting this property, we model the proposed manipulator using Cosserat rod theory~\cite{armanini2023soft}.

A Cosserat rod is a soft body capable of undergoing both rotational and translational deformations (i.e., bending, torsion, axial stretching, and shear), expressed by the strain vector
\begin{equation}
    \xi(X,t) : [0, L] \times [0,+\infty)
 \to \mathbb{R}^6 \, \text{,}
\end{equation}
where $L$ is the length of the rod, $t$ indicates the time, and $X$ represents the material curvilinear abscissa that parametrizes the distance along the central axis. 
This infinite-dimensional strain field can be parametrized as
\begin{equation}
    \xi(X,t) \approx \xi_n(X,t) = \xi_0(X)  + \sum_{i=1}^n q_i(t)b_{\mathrm{q},i}(X)
    \label{eq:xi}\, \text{,}
\end{equation}
where $\xi_n(X,t)$ is the approximated strain field, $\xi_0(X)$ is the reference strain field, and with
\begin{equation}
    B_{\mathrm{q}} (X) = \begin{bmatrix}
        b_{\mathrm{q},1}(X) & ... &b_{\mathrm{q},n}(X)
    \end{bmatrix}  \in \mathbb{R}^{6 \times n}\, \text{,}
\end{equation}
\begin{equation}
q (t)= \begin{bmatrix}
        q_1 (t) \\ \vdots \\q_n (t)
    \end{bmatrix}\in \mathbb{R}^{n}\, \text{,}
\end{equation}
being $B_{\mathrm{q}}(X)$ a matrix function whose columns form the basis for the strain field, $q(t)$ the generalized coordinates vector in that basis, and $n$ the resulting number of degrees of freedom. It is relevant to underline that for $n \to \infty$,~Eq.~\ref{eq:xi} is an exact equality. Since the strain basis can be a generic function of $X$, this approach is referred to as Variable Strain parametrization.

This procedure enables to study the dynamics of the soft robot with the well-known second-order differential equation 
\begin{equation}
M(q)\ddot{q} + (C(q,\dot{q}) + D(q))\dot{q} + Kq + F(q) = A(q)u \, \text{,}  \label{eq:dynamic_eq}
\end{equation}
where $M \in \mathbb{R}^{n \times n}$ is the generalized mass matrix, $C \in \mathbb{R}^{n \times n}$ the generalized Coriolis matrix, $D \in \mathbb{R}^{n \times n}$ the generalized damping matrix, $K \in \mathbb{R}^{n \times n}$ the generalized stiffness matrix,  $A \in \mathbb{R}^{n \times n_\mathrm{a}}$ the generalized actuation matrix, $F \in \mathbb{R}^{n}$ the vector of generalized external forces (e.g. gravity), and $u \in \mathbb{R}^{n_\mathrm{a}}$ the vector of applied actuation forces. The analytical expression of each matrix is detailed in the Supplementary Material and Methods. 

For thread-actuated systems, the actuation wrench exerted by the internal tendon (or chamber) on the central axis of the rod~\cite{renda_geometric_2020} is provided by
\begin{equation}
    \mathcal{F}_{\mathrm{a},k} =  \begin{bmatrix}
    {d}_k(X) \times t_{k}(X,q) \\
    t_{k}(X,q)
    \end{bmatrix} u_k = b_{\mathrm{\tau},k}(X,q) u_k\, \text{,}
\end{equation}
where 
$d_{k}(X) \in \mathbb{R}^3$ represents
the distance from the backbone to the $k$-th tendon, and $t_{k}(X) \in \mathbb{R}^3$ 
is the unit vector tangent to the actuator path. 

The contributions of all the internal tendons define the actuation basis
\begin{equation}
    B_{\mathrm{\tau}}(X,q) = \begin{bmatrix}
        b_{\mathrm{\tau},1}(X,q) & ... &b_{\mathrm{\tau},n_\mathrm{a}}(X,q)
    \end{bmatrix}  \in \mathbb{R}^{6\times n_\mathrm{a}}\, \text{.}
\end{equation}
Friction between the tendons and the manipulator alters how the applied pulling forces are transmitted to the system dynamics, thereby modifying the resulting actuation basis.
Taking inspiration from the work proposed in~\cite{rone2013continuum}, where friction effects are analyzed in case of rigid-body discretization, we aim to identify a friction model that explicitly accounts for the shape of the path followed by the tendons. As the soft arm changes configuration, the embodied tendon geometry deforms accordingly, thereby affecting the resulting frictional interactions. Moreover, frictional effects accumulate with increasing distance from the robot base, as more contact points between the tendons and the arm are encountered along the path.

To account for both these aspects, friction can be included by considering a new actuation basis whose columns are defined as follows:
\begin{equation}
    \bar{b}_{\mathrm{\tau},k}(X,q) = {b}_{\mathrm{\tau},k}(X,q) e^{-\mu_{\mathrm{S},k}(X,q)}\, \text{,}
\label{eq:act_matrix_eq_withfriction}
\end{equation}
with 
\begin{equation}
    \mu_{\mathrm{S},k}(X,q) = \bar{\mu}_{\phi,k} \int_0^X \phi_{k} (q,\eta)\, d\eta
    \label{eq:fric}\, \text{,}
\end{equation}
being $\bar{\mu}_{\phi,k}$ the friction coefficient for the $k$-th tendon, and $ \phi_{k} (q,X)$ the contribution arising from variations in the normalized cable force direction up to the abscissa $X$. This representation naturally incorporates the actual configuration of the robot, as the cable force direction changes according to $q$. Additionally, the minus sign of the exponential term in Eq.~\ref{eq:act_matrix_eq_withfriction} ensures higher friction effects (i.e., a reduction of the actuation basis magnitude) for increasing values of $X$. 

Finally, by projecting the new actuation basis $\bar{B}_{\mathrm{\tau}}(X,q)$ onto the
space of generalized coordinates and integrating over the length of the soft robot, the system's actuation matrix is obtained as
\begin{equation}
    A(q) = \int_{0}^{L} B_{\mathrm{q}}^T(X) \bar{B}_{\mathrm{\tau}}(X,q) \, dX
\label{eq::act_matrix_eq}\, \text{.}
\end{equation}
The detailed derivation, along with improvements in model accuracy with the new friction model, is further discussed in the Supplementary Materials and Methods.

The physical parameters of Eq.~\ref{eq:dynamic_eq} as well as the friction coefficient of Eq.~\ref{eq:fric} can be accurately estimated by suitable optimization problems, as further discussed in Supplementary Materials and Methods.

\subsection*{Low-level collocated control for dynamic shape regulation} 
Even when analytical reduced-order models are employed, soft robots typically exhibit underactuation, as the number of degrees of freedom $n$ far exceeds the number of actuators $n_\mathrm{a}$. To address this challenge, \cite{Pustina_coll2024} proposed a modeling strategy that implicitly accounts for underactuation in the control of Lagrangian systems. Adopting a similar approach within the VS modeling framework, the control design can be decoupled from the underactuation constraint by recasting the system in its \emph{Collocated form}. This is accomplished by introducing the coordinate transformation $\theta= \left[\theta_\mathrm{a}^T,\theta_\mathrm{u}^T \right]^T= h(q) \in \mathbb{R}^{n}$ which ensures that each of the first $n_\mathrm{a}$ equations of motion in~Eq.~\ref{eq:dynamic_eq} is influenced by one, and only one, independent actuator input:
\begin{equation}
\label{eq:sysdyn_theta}
\underbrace{
\begin{bmatrix}
M_{\theta_{\mathrm{a,a}}} & M_{\theta_{\mathrm{a,u}}} \\
M_{\theta_{\mathrm{u,a}}} & M_{\theta_{\mathrm{u,u}}}
\end{bmatrix}
}_{M_\theta}
\begin{bmatrix}
\ddot{\theta}_{\mathrm{a}} \\
\ddot{\theta}_{\mathrm{u}}
\end{bmatrix}
+ 
\Biggg( 
\underbrace{
\begin{bmatrix}
C_{\theta_{\mathrm{a,a}}} & C_{\theta_{\mathrm{a,u}}} \\
C_{\theta_{\mathrm{u,a}}} & C_{\theta_{\mathrm{u,u}}}
\end{bmatrix}
}_{C_\theta}
+ 
\underbrace{
\begin{bmatrix}
D_{\theta_{\mathrm{a,a}}} & D_{\theta_{\mathrm{a,u}}} \\
D_{\theta_{\mathrm{u,a}}} & D_{\theta_{\mathrm{u,u}}}
\end{bmatrix}
}_{D_\theta}
\Biggg) 
\begin{bmatrix}
\dot{\theta}_{\mathrm{a}} \\
\dot{\theta}_{\mathrm{u}}
\end{bmatrix}
+
\underbrace{
\begin{bmatrix}
K_{\theta_{\mathrm{a}}} \\
K_{\theta_{\mathrm{u}}}
\end{bmatrix}
}_{K_\theta}
+
\underbrace{
\begin{bmatrix}
F_{\theta_{\mathrm{a}}} \\
F_{\theta_{\mathrm{u}}}
\end{bmatrix}
}_{F_\theta}
=
\underbrace{
\begin{bmatrix}
 I_{n_\mathrm{a}} \\
 0_{(n-n_\mathrm{a}) \times n_\mathrm{a}}
\end{bmatrix}
}_{A_\theta}
u \, \text{,}
\end{equation}
where $\theta_\mathrm{u} \in \mathbb{R}^{n-n_{\mathrm{a}}}$ and $\theta_\mathrm{a} \in \mathbb{R}^{n_{\mathrm{a}}}$ denote the unactuated and actuation coordinates, respectively, with $\theta_\mathrm{a}$ satisfying the integrability condition 
\begin{equation}
    \dot{\theta}_\mathrm{a} = A^T(q) \dot{q}\, \text{.}
    \label{eq:thetadot_qdot}
\end{equation}

The system can now be controlled in this new set of coordinates to achieve a desired configuration $q_\mathrm{d}$, with underactuation addressed implicitly by mapping this reference to the corresponding desired actuation coordinates $\theta_\mathrm{a,d}$.

At this stage, the goal is to find a feedback controller $u(\theta,\dot{\theta})$ such that $\theta_\mathrm{a}$ converges asymptotically to an arbitrary desired value $\theta_\mathrm{a,d}$, i.e.
\begin{equation}
   \lim_{t \to \infty} {\theta}_\mathrm{a,d}-\theta_\mathrm{a}(t) = 0_{n_\mathrm{a}} \, \text{.}
\end{equation}
To achieve this, the following P-satI-D+ controller is employed
    \begin{equation}   
u = \Gamma_\mathrm{P}(\theta_{\mathrm{a,d}}-\theta_{\mathrm{a}})-\Gamma_\mathrm{D}\dot{\theta}_{\mathrm{a}} + \Gamma_\mathrm{I} \int_{0}^{t} {\mathrm{tanh}}(\theta_{\mathrm{a,d}}-\theta_{\mathrm{a}}(z))\, dz + (K_{\theta_\mathrm{a}}(q_{\mathrm{d}})+F_{\theta_\mathrm{a}}(q_{\mathrm{d}}))
\label{eq:PID+}
\end{equation}
with $\Gamma_\mathrm{P} >0,\Gamma_\mathrm{D}\ge0,\Gamma_\mathrm{I}>0 \in \mathbb{R}^{n_\mathrm{a} \times n_\mathrm{a}}$ being the gain matrices.
The asymptotic convergence of ${\theta}_\mathrm{a}$ to $\theta_\mathrm{a,d}$ under this control law has been previously assessed in~\cite{pustinaPSatID}.

For tendon-actuated systems without friction modeling, the actuation coordinates coincide with the tendon lengths, i.e.
\begin{equation}        
 \theta_{\mathrm{a},{k}}(q) = L_{\mathrm{c},{k}}(q) = \int_{0}^{L} t_{k}^T(X,q)  
\left[\hat{\xi}\,d_{k}(X) + d_{k}'(X)\right]_3 \, dX \, \text{,}
\end{equation} 
where $\hat{(\cdot)}$ denotes the Lie algebra $\mathfrak{se}(3)$ mapping and $(\cdot)'$ the spatial derivative with respect to $X$. This equality implies that the PID term depends exclusively on tendon length measurements rather than on the robot configuration $q$, which lacks a direct physical interpretation and therefore cannot be measured directly. Additionally, the feedforward compensation of gravity and elastic terms speeds up convergence while requiring only the assigned desired configuration.

However, when friction effects are included, the actuation matrix changes - relying on $\bar{B}_\tau$ instead of $B_\tau$ - breaking the equality between actuation coordinates and tendon lengths.
Specifically, for the configuration-dependent friction model, the actuation coordinate for the $k$-th tendon is redefined as 
\begin{equation} \theta_{\mathrm{a},{k}}(q) = \int_{0}^{L} t_{k}^T(X,q)  e^{-{\mu}_{\mathrm{S},k}(X,q)}
\left[\hat{\xi}\,d_{k}(X) + d_{k}'(X)\right]_3 \, dX  \text{,} 
\end{equation}
which, under a quasi-static assumption, satisfies the passive output definition in Eq.~\ref{eq:thetadot_qdot} (see Supplementary Materials and Methods).
 
The exact reconstruction of $\theta_{\mathrm{a},k}$ requires knowledge of both the generalized coordinates $q$ and the tendon lengths $L_{\mathrm{c},k}$. However, $q$ cannot be directly measured when adopting Variable Strain models. Simulation results show that, in practice, the actuation coordinates depend primarily on the tendon lengths alone (see Supplementary Materials and Methods). This allows us to approximate $\theta_{\mathrm{a}}(q,L_{\mathrm{c}}) \approx \bar{\theta}_{\mathrm{a}}(L_{\mathrm{c}})$, enabling the controller to rely solely on tendon length feedback.


\subsection*{Closed-Loop Inverse Kinematics for underactuated systems}
To control the robot in the operational space, it is necessary to establish a relationship between the generalized and task-space coordinates. In fully actuated systems, such as rigid robots, this procedure is typically implemented using a CLIK algorithm that actively compensates for modeling uncertainties and external disturbances, ensuring error convergence and robustness. However, extending this approach to underactuated soft robots is challenging, as not all configurations can be attained at steady state via control actions. The IK problem for underactuated systems can be formalized as follows: 
\\find a mapping $c_{\mathrm{IK}}:\mathbb{R}^n \times \mathbb{R}^m \to \mathbb{R}^n$ such that
\begin{equation}
    \dot{q}=c_{\mathrm{IK}}(q,x_\mathrm{d}),\text{ s.t. } \lim_{t \to \infty} h_\mathrm{g}(q(t))=x_\mathrm{d} \land \lim_{t \to \infty} q(t) \in \epsilon
   \label{eq:clik_gen}
\end{equation}
where $\epsilon$ defines the attainable equilibria set
\begin{equation}
    \epsilon = \left\{ q \in \mathbb{R}^n | \exists u \in \mathbb{R}^{n_\mathrm{a}} \text{ s.t. } G(q)=A(q)u \right\} \subset \mathbb{R}^n
    \label{eq:eq_set}\, \text{.}
\end{equation}
 Here, $G(q)$ accounts for both elastic and gravitational forces, $x_\mathrm{d} \in \mathbb{R}^m$ is the desired position, and $x=h_\mathrm{g}(q) \in \mathbb{R}^m$ is the actual one in the operational space ($m \leq 6$).

To address this challenge, this work adapts the CLIK algorithm with convergence proof for underactuated Lagrangian systems proposed in~\cite{santina_clik25} within the VS modeling framework. By leveraging the collocated form of the dynamic system, the algorithm imposes a squared mapping that enables to control as many task-space dimension as the number of actuation coordinates, i.e. $m=n_\mathrm{a}$. In this way, convergence to the desired pose is achieved by closing the following loop:
\begin{equation}
    \dot{\theta} = \bar{P}_\epsilon ({\theta}_\mathrm{a},{\theta}_\mathrm{u})\Gamma_\mathrm{Z}(x_\mathrm{d}-x) - \bar{P}_\mathrm{Z}({\theta}_a,{\theta}_u)\Gamma_\mathrm{\epsilon} G_{\theta _\mathrm{u}}({\theta}_\mathrm{a},{\theta}_\mathrm{u})
    \label{eq:click2}\, \text{,}
\end{equation}
where $\Gamma_\mathrm{Z} \in \mathbb{R}^{m \times m}$ and $\Gamma_\mathrm{\epsilon} \in \mathbb{R}^{n-n_\mathrm{a} \times n-n_\mathrm{a}}$ are two gain matrices, $\bar{P}_\mathrm{\epsilon}\in \mathbb{R}^{n \times m}$ and $\bar{P}_\mathrm{Z}\in \mathbb{R}^{n \times {n-n_\mathrm{a}}}$ are two projectors used to ensure that the two feedback actions are decoupled and do not disturb each other, and $G_{\theta _\mathrm{u}}$ is the contribution of the elastic and gravitational forces on the unactuated coordinates $\theta_\mathrm{u}$. As can be observed, this equation consists of two terms: the first implements a standard kinematic inversion driven by the convergence of the task-space error $e = x_\mathrm{d} - x$, while the second enforces convergence toward a static equilibrium, which is achieved when the following condition is satisfied:
\begin{equation}
    G_{\theta _\mathrm{u}}= K_{\theta _\mathrm{u}}+F_{\theta _\mathrm{u}} = 0_{n-n_\mathrm{a} }
    \label{eq:G_u}\, \text{,}
\end{equation}
in accordance with~Eq.~\ref{eq:sysdyn_theta}.

Once the generalized velocity in the collocated form $\dot{\theta}$ is obtained, the corresponding velocity in the original form $\dot{q}$ is retrieved by inverting the Jacobian between the two representations:
\begin{equation}
   \dot{q}= J_\mathrm{h}^{-1} \dot{\theta}\, \text{,}
\end{equation}
with 
\begin{equation}
   J_\mathrm{h}= \begin{bmatrix}
       \multicolumn{2}{c}{A^T}  \\
       0_{(n-n_\mathrm{a})\times n_\mathrm{a}} & I_{n-n_\mathrm{a}} 
   \end{bmatrix}\, \text{.}
\end{equation}
The desired configuration $q_\mathrm{d}$ is then obtained by integrating $\dot{q}$ over time. It is worth emphasizing that this mapping is exact and introduces no approximation, since the inversion of $J_\mathrm{h}$ is well-defined throughout the motion and all quantities involved correspond to known desired reference values.

Notably, for the precomputed CLIK algorithm, the robot pose $x$ is directly retrieved from the prediction of the model, whereas in the online version it is measured by the motion capture system.

\clearpage 

%
\bibliography{mybib} %

\begin{thebibliography}{10}
\providecommand{\url}[1]{\texttt{#1}}
\expandafter\ifx\csname urlstyle\endcsname\relax
  \providecommand{\doi}[1]{doi:\discretionary{}{}{}#1}\else
  \providecommand{\doi}{doi:\discretionary{}{}{}\begingroup \urlstyle{rm}\Url}\fi

\bibitem{rus2015design}
D.~Rus, M.~T. Tolley, Design, fabrication and control of soft robots. \emph{Nature} \textbf{521}, 467--475 (2015).

\bibitem{whitesides2018soft}
G.~M. Whitesides, Soft robotics. \emph{Angewandte Chemie International Edition} \textbf{57}, 4258--4273 (2018).

\bibitem{xie2023octopus}
Z.~Xie, \emph{et~al.}, Octopus-inspired sensorized soft arm for environmental interaction. \emph{Science Robotics} \textbf{8} (2023).

\bibitem{yue2025embodying}
T.~Yue, \emph{et~al.}, Embodying soft robots with octopus-inspired hierarchical suction intelligence. \emph{Science Robotics} \textbf{10} (2025).

\bibitem{laschi2016soft}
C.~Laschi, B.~Mazzolai, M.~Cianchetti, Soft robotics: Technologies and systems pushing the boundaries of robot abilities. \emph{Science robotics} \textbf{1} (2016).

\bibitem{wang2024sensing}
P.~Wang, \emph{et~al.}, Sensing expectation enables simultaneous proprioception and contact detection in an intelligent soft continuum robot. \emph{Nature Communications} \textbf{15} (2024).

\bibitem{bang2024bioinspired}
J.~Bang, \emph{et~al.}, Bioinspired electronics for intelligent soft robots. \emph{Nature Reviews Electrical Engineering} \textbf{1}, 597--613 (2024).

\bibitem{thuruthel2018model}
T.~G. Thuruthel, E.~Falotico, F.~Renda, C.~Laschi, Model-based reinforcement learning for closed-loop dynamic control of soft robotic manipulators. \emph{IEEE Transactions on Robotics} \textbf{35}, 124--134 (2018).

\bibitem{santina_planarSR2020}
C.~Della~Santina, R.~K. Katzschmann, A.~Bicchi, D.~Rus, Model-based dynamic feedback control of a planar soft robot: trajectory tracking and interaction with the environment. \emph{The International Journal of Robotics Research} \textbf{39}, 490--513 (2020).

\bibitem{centurelli2022}
A.~Centurelli, \emph{et~al.}, Closed-Loop Dynamic Control of a Soft Manipulator Using Deep Reinforcement Learning. \emph{IEEE Robotics and Automation Letters} \textbf{7}, 4741--4748 (2022).

\bibitem{fischer2023dynamic}
O.~Fischer, Y.~Toshimitsu, A.~Kazemipour, R.~K. Katzschmann, Dynamic Task Space Control Enables Soft Manipulators to Perform Real-World Tasks. \emph{Advanced Intelligent Systems} \textbf{5} (2023).

\bibitem{bruder2025koopman}
D.~Bruder, D.~Bombara, R.~J. Wood, A Koopman-based residual modeling approach for the control of a soft robot arm. \emph{The International journal of robotics research} \textbf{44}, 388--406 (2025).

\bibitem{azizkhani2025dynamic}
M.~Azizkhani, S.~Kousik, Y.~Chen, Dynamic Task Space Control of Redundant Pneumatically Actuated Soft Robot. \emph{IEEE Robotics and Automation Letters}  (2025).

\bibitem{Rus_soft2026}
Z.~Tang, \emph{et~al.}, A general soft robotic controller inspired by neuronal structural and plastic synapses that adapts to diverse arms, tasks, and perturbations. \emph{Science Advances} \textbf{12} (2026).

\bibitem{li2023piecewise}
H.~Li, L.~Xun, G.~Zheng, Piecewise linear strain Cosserat model for soft slender manipulator. \emph{IEEE transactions on robotics} \textbf{39}, 2342--2359 (2023).

\bibitem{mathew_reduced_2025}
A.~T. Mathew, D.~Feliu-Talegon, A.~Y. Alkayas, F.~Boyer, F.~Renda, Reduced order modeling of hybrid soft-rigid robots using global, local, and state-dependent strain parameterization. \emph{The International Journal of Robotics Research} \textbf{44}, 129--154 (2025).

\bibitem{della2023model}
C.~Della~Santina, C.~Duriez, D.~Rus, Model-based control of soft robots: A survey of the state of the art and open challenges. \emph{IEEE Control Systems Magazine} \textbf{43}, 30--65 (2023).

\bibitem{Pustina_coll2024}
P.~Pustina, C.~D. Santina, F.~Boyer, A.~De~Luca, F.~Renda, Input Decoupling of Lagrangian Systems via Coordinate Transformation: General Characterization and Its Application to Soft Robotics. \emph{IEEE Transactions on Robotics} \textbf{40}, 2098--2110 (2024).

\bibitem{santina_clik25}
C.~Della~Santina, Pushing the boundaries of actuators-to-task kinematic inversion: from fully actuated to underactuated (soft) robots. \emph{Authorea Preprints}  (2025).

\bibitem{haggerty2023control}
D.~A. Haggerty, \emph{et~al.}, Control of soft robots with inertial dynamics. \emph{Science robotics} \textbf{8} (2023).

\bibitem{bruder2020data}
D.~Bruder, X.~Fu, R.~B. Gillespie, C.~D. Remy, R.~Vasudevan, Data-driven control of soft robots using Koopman operator theory. \emph{IEEE transactions on robotics} \textbf{37}, 948--961 (2020).

\bibitem{franka_research3_datasheet_2025}
{Franka Robotics GmbH}, \emph{Franka Research 3 Datasheet, Document R02212, Version 2.4}, Tech. rep. (2025), \url{https://franka.de/hubfs/Datasheet_FrankaResearch3_R02212_2.4.pdf}.

\bibitem{franzesecal}
G.~Franzese, M.~Spahn, J.~Kober, J.~Alonso-Mora, C.~Della~Santina, Accurate and Affordable Cobot Calibration Without External Measurement Devices. \emph{Nature Communications Engineering}  (2026).

\bibitem{armanini2023soft}
C.~Armanini, F.~Boyer, A.~T. Mathew, C.~Duriez, F.~Renda, Soft robots modeling: A structured overview. \emph{IEEE Transactions on Robotics} \textbf{39}, 1728--1748 (2023).

\bibitem{renda_geometric_2020}
F.~Renda, C.~Armanini, V.~Lebastard, F.~Candelier, F.~Boyer, A Geometric Variable-Strain Approach for Static Modeling of Soft Manipulators With Tendon and Fluidic Actuation. \emph{{IEEE} Robotics and Automation Letters} \textbf{5}, 4006--4013 (2020).

\bibitem{rone2013continuum}
W.~S. Rone, P.~Ben-Tzvi, Continuum robot dynamics utilizing the principle of virtual power. \emph{IEEE Transactions on Robotics} \textbf{30}, 275--287 (2013).

\bibitem{pustinaPSatID}
P.~Pustina, P.~Borja, C.~D. Santina, A.~De~Luca, P-satI-D Shape Regulation of Soft Robots. \emph{IEEE Robotics and Automation Letters} \textbf{8}, 1--8 (2023).

\end{thebibliography}
\bibliographystyle{sciencemag}

%
%
%
%
%
%

\newpage

\end{document}